\documentclass[runningheads]{llncs}
\usepackage{graphicx}
\usepackage{CJK}
\usepackage{amsmath}
\usepackage{amsfonts}
\usepackage{longtable}
\usepackage{multirow}
\usepackage[colorlinks,linkcolor=blue,anchorcolor=blue,citecolor=blue]{hyperref}

\newcommand{\IntEnt}[1]{\left[\!\left[#1\right]\!\right]}
\usepackage{color}
\newcommand{\modif}[1]{{\color{black}{#1}}}

\newcommand{\R}{\mathbb{R}}

\usepackage[ruled,linesnumbered]{algorithm2e}
\usepackage{setspace}

\begin{document}
\title{Sentence Compression via DC Programming Approach\thanks{The research is partially funded by the Natural Science Foundation of China (Grant No: 11601327) and by the Key Construction National ``$985$" Program of China (Grant No: WF220426001).}}
%
\author{Yi-Shuai Niu\inst{1,2}\orcidID{0000-0002-9993-3681} \and
Xi-Wei Hu\inst{2} \and Yu You\inst{1} \\
Faouzi Mohamed Benammour\inst{1} \and Hu Zhang\inst{1}}
\authorrunning{Y.S. Niu et al.}
%
\institute{School of Mathematical Sciences, Shanghai Jiao Tong University, Shanghai, China
\email{niuyishuai@sjtu.edu.cn}\\
\and
SJTU-Paristech Elite Institute of Technology, Shanghai Jiao Tong University, Shanghai, China}
\maketitle              
\begin{abstract}

Sentence compression is an important problem in natural language processing. In this paper, we firstly establish a new sentence compression model based on the probability model and the parse tree model. Our sentence compression model is equivalent to an integer linear program (ILP) which can both guarantee the syntax correctness of the compression and save the main meaning. We propose using a DC (Difference of convex) programming approach (DCA) for finding local optimal solution of our model. Combing DCA with a parallel-branch-and-bound framework, we can find global optimal solution. Numerical results demonstrate the good quality of our sentence compression model and the excellent performance of our proposed solution algorithm.
\keywords{Sentence Compression  \and Probability Model \and Parse Tree Model \and DCA \and Parallel-Branch-and-Bound }
\end{abstract}

\section{Introduction}\label{sec:introduction}

The recent years have been known by the quick evolution of the artificial intelligence (AI) technologies, and the sentence compression problems attracted the attention of researchers due to the necessity of dealing with a huge amount of natural language information in a very short response time. The general idea of sentence compression is to make a summary with shorter sentences containing the most important information while maintaining grammatical rules. Nowadays, there are various technologies involving sentence compression as: text summarization, search engine and question answering etc. Sentence compression will be a key technology in future human-AI interaction systems.

There are various models proposed for sentence compression. The paper of Jing \cite{paper_jing_2000} could be one of the first works addressed on this topic with many rewriting operations as deletion, reordering, substitution, and insertion. This approach is realized based on multiple knowledge resources (such as WordNet and parallel corpora) to find the pats that can not be removed if they are detected to be grammatically necessary by using some simple rules. Later, Knight and Marcu investigated discriminative models\cite{paper_knight_2002}. They proposed a decision-tree model to find the intended words through a tree rewriting process, and a noisy-channel model to construct a compressed sentence from some scrambled words based on the probability of mistakes. MacDonald \cite{paper_mcdonald_2006} presented a sentence compression model using a discriminative large margin algorithm. He ranks each candidate compression using a scoring function based on the Ziff-Davis corpus using a Viterbi-like algorithm. The model has a rich feature set defined over compression bigrams including parts of speech, parse trees, and dependency information, without using a synchronous grammar. Clarke and Lapata \cite{paper_clarke_2008} reformulated McDonald's model in the context of integer linear programming (ILP) and extended with constraints ensuring that the compressed output is grammatically and semantically well formed. The corresponding ILP model is solving in using the branch-and-bound algorithm.

In this paper, we will propose a new sentence compression model to both guarantee the grammatical rules and preserve main meaning. The main contributions in this work are: (1) Taking advantages of \emph{Parse tree model} and \emph{Probability model}, we hybridize them to build a new model that can be formulated as an ILP. Using the Parse tree model, we can extract the sentence truck, then fix the corresponding integer variables in the Probability model to derive a simplified ILP with improved quality of the compressed result. (2) We propose to use a DC programming approach called PDCABB (an hybrid algorithm combing DCA with a parallel branch-and-bound framework) developed by Niu in \cite{paper_niu_2018} for solving our sentence compression model. This approach can often provide a high quality optimal solution in a very short time.

The paper is organized as follows: The Section \ref{sec:sentence_compression_models} is dedicated to establish hybrid sentence compression model. In Section \ref{sec:solvethemodel}, we will present DC programming approach for solving ILP. The numerical simulations and the experimental setup will be reported in Section \ref{sec:experimental_result}. Some conclusions and future works will be discussed in the last section.

%
\section{Hybrid Sentence Compression Model}\label{sec:sentence_compression_models}
Our sentence compression model is based on an Integer Linear Programming (ILP) probability model \cite{paper_clarke_2008}, and a parsing tree model. In this section, we will give a brief introduction of the two models, and propose our new hybrid model.

\subsection{ILP Probability Model}\label{subsec:ILPmodel}

Let \textbf{x} = $\{x_1, x_2, \ldots, x_n\}$ be a sentence with $n \geq 2$ words.\footnote{Punctuation is also deemed as word.} We add $x_0$=`start' as the start token and $x_{n+1}$=`end' as the end token.  

The sentence compression is to choose a subset of words in $\textbf{x}$ for maximizing its probability to be a sentence under some restrictions to the allowable trigram combinations. This probability model can be described as an ILP as follows:
\smallskip

\noindent\textbf{Decision variables:}
We introduce the binary \emph{decision variables} $\delta_i$,
\begin{math}
i\in \IntEnt{1,n}\end{math}\footnote{$\IntEnt{m,n}$ with $m\leq n$ stands for the set of integers between $m$ and $n$.}
for each word $x_i$ as: $\delta_i=1$ if $x_i$ is in a compression and $0$ otherwise. In order to take context information into consideration, we introduce the \emph{context variables} $(\alpha,\beta,\gamma)$ such that: $\forall i \in \IntEnt{1,n}$, we set $\alpha_i=1$ if $x_i$ starts a compression and $0$ otherwise; $\forall i \in \IntEnt{0,n-1}, j \in \IntEnt{i+1,n}$, we set $\beta_{ij}=1$ if the sequence $x_i$, $x_j$ ends a compression and $0$ otherwise; and $\forall i \in \IntEnt{0,n-2}, j \in \IntEnt{i+1,n-1}, k \in \IntEnt{j+1,n}$, we set $\gamma_{ijk}=1$ if sequence $x_i$, $x_j$, $x_k$ is in a compression and $0$ otherwise. There are totally $\frac{n^3+3n^2+14n}{6}$ binary variables for $(\delta,\alpha,\beta,\gamma)$.

\smallskip
\noindent\textbf{Objective function:} The objective function is to maximize the probability of the compression computed by:
\begin{eqnarray}
f(\alpha,\beta,\gamma) &=& \sum_{i=1}^n\alpha_i P\left(x_i|\text{start}\right)+
\sum_{i=1}^{n-2}\sum_{j=i+1}^{n-1}\sum_{k=j+1}^n\gamma_{ijk} P\left(x_k|x_i, x_j\right) \nonumber\\
&& + \sum_{i=0}^{n-1}\sum_{j=i+1}^n\beta_{ij} P\left(\text{end}|x_i, x_j\right)\nonumber
\end{eqnarray}
where $P\left(x_i|\text{start}\right)$ stands for the probability of a sentence starting with $x_i$, $P\left(x_k|x_i, x_j\right)$ denotes the probability that $x_i,x_j,x_k$ successively occurs in a sentence, and $P\left(\text{end}|x_i, x_j\right)$ means the probability that $x_i,x_j$ ends a sentence. The probability $P\left(x_i|\text{start}\right)$ is computed by bigram model, and the others are computed by trigram model based on some corpora.


\smallskip
\noindent\textbf{Constraints:}
The following sequential constraints will be introduced to restrict the possible trigram combinations:

\noindent{\bf Constraint 1} Exactly one word can begin a sentence.
	\begin{equation}\label{cons:1}
	\sum_{i=1}^n\alpha_i = 1.
	\end{equation}
	
\noindent{\bf Constraint 2} If a word is included in a compression, it must either start the sentence, or be preceded by two other words, or be preceded by the `start' token and one other word.
	\begin{equation}\label{cons:2}
	\delta_k-\alpha_k-\sum_{i=0}^{k-2}\sum_{j=1}^{k-1}\gamma_{ijk} = 0, \forall k \in\IntEnt{1,n}.
	\end{equation}
	
\noindent{\bf Constraint 3} If a word is included in a compression, it must either be preceded by one word and followed by another, or be preceded by one word and end the sentence.
	\begin{equation}\label{cons:3}
	\delta_j-\sum_{i=0}^{j-1}\sum_{k=j+1}^{n}\gamma_{ijk}-\sum_{i=0}^{j-1}\beta_{ij} = 0, \forall j \in \IntEnt{1,n}.
	\end{equation}
	
\noindent{\bf Constraint 4} If a word is in a compression, it must either be followed by two words, or be followed by one word and end the sentence. 
	\begin{equation}\label{cons:4}
	\delta_i-\sum_{j=i+1}^{n-1}\sum_{k=j+1}^{n}\gamma_{ijk}-\sum_{j=i+1}^{n}\beta_{ij}
	-\sum_{h=0}^{i-1}\beta_{hi} = 0, \forall i \in \IntEnt{1,n}.
	\end{equation}

\noindent{\bf Constraint 5} Exactly one word pair can end the sentence.
	\begin{equation}\label{cons:5}
	\sum_{i=1}^{n-1}\sum_{j=i+1}^n\beta_{ij} = 1.
	\end{equation}

\noindent{\bf Constraint 6} The length of a compression should be bounded.
	\begin{equation}\label{cons:6}
	\underline{l}\leq  \sum_{i=1}^n\delta_i \leq \bar{l}.
	\end{equation}
	with given lower and upper bounds of the compression $\underline{l}$ and $\bar{l}$. 
	
\noindent{\bf Constraint 7} The introducing term for preposition phrase (PP) or subordinate clause (SBAR) must be included in the compression if any word of the phrase is included. Otherwise, the phrase should be entirely removed. Let us denote $I_i = \{j: x_j\in \text{PP/SBAR}, j\neq i\}$ the index set of the words included in PP/SBAR leading by the introducing term $x_i$, then
	\begin{equation}\label{cons:7}
	\sum_{j\in I_i}\delta_j \geq \delta_i, \delta_i \geq \delta_j, \forall j\in I_i.
	\end{equation}

\noindent\textbf{ILP probability model:} The optimization model for sentence compression is summarized as a binary linear program as:
	\begin{equation}\label{prob:ILP}
	\max\{f(\alpha,\beta,\gamma): (\ref{cons:1})-(\ref{cons:7}), (\alpha,\beta,\gamma,\delta)\in\{0,1\}^{\frac{n^3+3n^2+14n}{6}} \}.
	\end{equation}
	with $O(n^3)$ binary variables and $O(n)$ linear constraints. 

The advantage of this model is that its solution will provide a compression with maximal probability based on the trigram model. However, there is no information about syntactic structures of the target sentence, so it is possible to generate ungrammatical sentences. In order to overcome this disadvantage, we propose to combine it with the parse tree model presented below.

\subsection{Parse Tree Model}\label{subsec:parsetreemodel}

A parse tree is an ordered, rooted tree which reflects the syntax of the input language based on some grammar rules (e.g. using CFG syntax-free grammar). For constructing a parse tree in practice, we can use a nature language processing toolkit NLTK \cite{software_NLTK} in Python. Based on NLTK, we have developed a CFG grammar generator which helps to generate automatically a CFG grammar based on a target sentence. A recursive descent parser can help to build a parse tree.
	
\begin{figure}[h!]
	\centering
	\includegraphics[width=6cm]{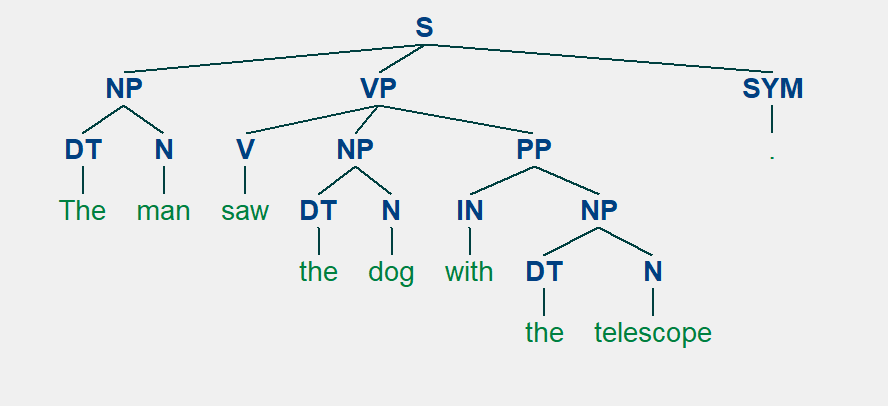}
	\caption{Parse tree example}\label{fig:1}
\end{figure}

For example, the sentence ``The man saw the dog with the telescope.'' can be parsed as in Figure \ref{fig:1}. It is observed that a higher level node in the parse tree indicates more important sentence components (e.g., the sentence S consists of a noun phrase NP, a verb phrase VP, and a symbol SYM), whereas a lower node tends to carry more semantic contents (e.g., the proposition phrase PP is consists of a preposition `with', and a noun phrase `the telescope'). Therefore, a parse tree presents the clear structure of a sentence in a logical way.

Sentence compression can be also considered as finding a subtree which remains grammatically correct and containing main meaning of the original sentence. Therefore, we can propose a procedure to delete some nodes in the parse tree. For instance, the sentence above can be compressed as ``The man saw the dog.'' by deleting the node PP.

\subsection{New Hybrid Model: ILP-Parse Tree Model}

Our proposed model for sentence compression, called ILP-Parse Tree Model (ILP-PT), is based on the combination of the two models described above. The ILP model will provide some candidates for compression with maximal probability, while the parse tree model helps to guarantee the grammar rules and keep the main meaning of the sentence. This combination is described as follows:

\noindent\textbf{Step 1 (Build ILP probability model):} Building the ILP model as in formulation (\ref{prob:ILP}) for the target sentence.

\noindent\textbf{Step 2 (Parse Sentence):} Building a parse tree as described in subsection \ref{subsec:parsetreemodel}.

\noindent\textbf{Step 3 (Fix variables for sentence trunk):} Identifying the sentence trunk in the parse tree and fixing the corresponding integer variables to be $1$ in ILP model. This step helps to extract the sentence trunk by keeping the main meaning of the original sentence while reducing the number of binary decision variables. 

More precisely, we will introduce for each node $N_i$ of the parse tree a label $s_{N_i}$ taking the values in $\left\{0, 1, 2\right\}$. A value $0$ represents the deletion of the node; $1$ represents the reservation of the node; whereas $2$ indicates that the node can either be deleted or be reserved. We set these labels as \emph{compression rules} for each CFG grammar to support any sentence type of any language. 

For the word $x_i$, we go through all its parent nodes till the root S. If the traversal path contains $0$, then $\delta_i=0$; else if the traversal path contains only $1$, then $\delta_i=1$; otherwise $\delta_i$ will be further determined by solving the ILP model. The sentence truck is composed by the words $x_i$ whose $\delta_i$ are fixed to $1$. Using this method, we can extract the sentence trunk and reduce the number of binary variables in ILP model. 

\modif{\noindent\textbf{Step 4 (Solve ILP):} Applying an ILP solution algorithm to solve the simplified ILP model derived in Step 3 and generate a compression. In the next section, we will introduce a DC programming approach for solving ILP.}

\section{DC Programming approach for solving ILP}\label{sec:solvethemodel}
\modif{Solving an ILP is in general NP-hard. A classical and most frequently used method is branch-and-bound algorithm as in \cite{paper_clarke_2008}. Gurobi \cite{software_gurobi} is currently one of the best ILP solvers, which is an efficient implementation of branch-and-bound combing various techniques such as presolve, cutting planes, heuristics and parallelism etc.

In this section, we will present a Difference of Convex (DC) programming approach, called DCA-Branch-and-Bound (DCABB), for solving this model. DCABB is initially designed for solving mixed-integer linear programming (MILP) proposed in \cite{paper_niu_2008}, and extended for solving mixed-integer nonlinear programming \cite{paper_niu_2010,paper_niu_2011} with various applications including scheduling \cite{paper_lethi_2009a}, network optimization \cite{paper_schleich_2012}, cryptography \cite{paper_lethi_2009c} and finance \cite{paper_lethi_2009b,paper_pham_2016} etc. This algorithm is based on continuous representation techniques for integer set, exact penalty theorem, DCA and Branch-and-Bound algorithms. Recently, the author developed a parallel branch-and-bound framework (called PDCABB) \cite{paper_niu_2018} in order to use the power of multiple CPU and GPU for improving the performance of DCABB. 

The ILP model can be stated in standard matrix form as: 
\begin{equation*}\label{$P$}
\min \{f(x):=c^{\top}x: x\in S\}
\tag{$P$}
\end{equation*}
where $S = \{x\in \{0,1\}^n: Ax = b\}$, $c\in \R^n$, $b \in \R^m$ and $A \in \R^{m\times n}$. Let us denote $K$ the linear relaxation of $S$ defined by $K = \{x\in [0,1]^n: Ax = b\}$. Thus, we have the relationship between $S$ and $K$ as $S = K \cap \{0,1\}^n.$

The linear relaxation of (\ref{$P$}) denoted by $R(P)$ is  
\begin{equation*}\label{$RP$}
\min \{f(x): x\in K\},
\end{equation*}
whose optimal value denoted by $l(P)$ is a lower bound of (\ref{$P$}).

The continuous representation technique for integer set $\{0,1\}^n$ consists of finding a continuous DC function\footnote{A function $f:\R^n\to \R$ is called DC if there exist two convex functions $g$ and $h$ (called DC components) such that $f=g-h$.} $p:\R^n\to \R$ such that $$\{0,1\}^n \equiv \{x: p(x)\leq 0 \}.$$

We often use the following functions for $p$ with their DC components:\\
\begin{tabular}[h]{l|l|l}		
		\hline
		function type & expression of $p$ & DC components of $p$\\
		\hline
		piecewise linear & $\sum_{i=1}^{n} \min\{x_i,1-x_i\}$ & \multirow{2}*{$g(x) = 0$, $h(x)=-p(x)$}\\
		\cline{1-2}
		quadratic & $\sum_{i=1}^{n}x_i(1-x_i)$ & \\
		\hline
		trigonometric & $\sum_{i=1}^{n}\sin^2(\pi x_i)$ & $g(x) = \pi^2 \|x\|^2$, $h(x) = g(x) - p(x)$\\
		\hline
\end{tabular}

Based on the exact penalty theorem \cite{LeThi_1999,LeThi_2012}, there exists a large enough parameter $t\geq 0$ such that the problem (\ref{$P$}) is equivalent to the problem (\ref{$Pt$}):
\begin{equation*}
\min \{ F_t(x):= f(x) + t p(x) : x \in K\} 
\label{$Pt$}.
\tag{$P^t$}
\end{equation*}
The objective function $F_t: \R^n\to \R$ in (\ref{$Pt$}) is also DC with DC components $g_t$ and $h_t$ defined as $g_t(x) = tg(x), h_t(x) = th(x) - f(x)$ where $g$ and $h$ are DC components of $p$. Thus the problem (\ref{$Pt$}) is a DC program which can be solved by DCA described in Algorithm \ref{algo:DCA}.

\begin{algorithm}[h!]
	\setstretch{0.3}
	\modif{
	\caption{DCA for (\ref{$Pt$})}
	\label{algo:DCA}
	\KwIn{Initial point $x^0\in \R^n$; large enough penalty parameter $t>0$; tolerance $\varepsilon_1,\varepsilon_2>0$.}
	\KwOut{Optimal solution $x^*$ and optimal value $f^*$;} 
	
	\textbf{Initialization:} Set $i=0$. 
	
	\textbf{Step 1:} Compute $y^i \in \partial h(x^i)$;
	
	\textbf{Step 2:} Solve $x^{i+1} \in \arg \min \{ g(x) - \langle x, y^i \rangle : x \in K \}$;
	
	\textbf{Step 3:} Stopping check: 	

	\SetKwIF{If}{ElseIf}{Else}{if}{then}{else if}{else}{end}
	\eIf{$\|x^{i+1}-x^{i}\| \leq \varepsilon_1$ or $|F_t(x^{i+1})-F_t(x^{i})| \leq \varepsilon_2$}
	{
		$x^* \leftarrow x^{i+1}$; $f^* \leftarrow F_t(x^{i+1})$; \Return;
	}{	
	$i\leftarrow i+1$; \textbf{Goto Step 1}.}
	
}	
\end{algorithm}
The symbol $\partial h (x^i)$ denotes the subdifferential of $h$ at $x^i$ which is fundamental in convex analysis. The subdifferential generalizes the derivative in the sense that $h$ is differentiable at $x^i$ if and only if $\partial h(x^i)$ reduces to the singleton $\{ \nabla h(x^i)\}$. 

Concerning on the choice of the penalty parameter $t$, we suggest using the following two methods: the first method is to take arbitrarily a large value for $t$; the second one is to increase $t$ by some ways in iterations of DCA (e.g., \cite{paper_niu_2010,paper_pham_2016}). Note that a smaller parameter $t$ yields a better DC decomposition \cite{paper_niu_2018a}.


Concerning on the numerical results given by DCA, it is often observed that DCA provides an integer solution which is also an upper bound solution for the problem (\ref{$P$}). Therefore, DCA is often proposed for upper bound algorithm in nonconvex optimization. More details about DCA and its convergence theorem can be found in \cite{LeThi_2005,LeThi_Home}. Combing DCA with a parallel-branch-and-bound algorithm (PDCABB) proposed in \cite{paper_niu_2018}, we can globally solve ILP. The PDCABB algorithm is described in Algorithm \ref{algo:PDCABB}. More details about this algorithm as the convergence theorem, branching strategies, parallel node selection strategies will be discussed in full-length paper.}

\begin{algorithm}[h!]
	\setstretch{0.3}
	\modif{
		\caption{PDCABB}
		\label{algo:PDCABB}
		\KwIn{Problem (\ref{$P$}); number of parallel workers $s$; tolerance $\varepsilon>0$;}
		\KwOut{Optimal solution $x_{opt}$ and optimal value $f_{opt}$;} 
		
		\textbf{Initialization:} $x_{opt}=null$; $f_{opt}=+\infty$.
		
		\textbf{Step 1: Root Operations}
		
		Solve $R(P)$ to obtain its optimal solution $x^*$ and set $LB\leftarrow l(P)$;
		
		\SetKwIF{If}{ElseIf}{Else}{if}{then}{else if}{else}{end}
		\uIf{$R(P)$ is infeasible}
		{
			
			\Return;
		}
		\ElseIf{$x^*\in S$}
		{
			$x_{opt} \leftarrow x^*$; $f_{opt} \leftarrow LB$; \Return;
		}
		
		Run DCA for (\ref{$Pt$}) from $x^*$ to get $\bar{x^*}$;
		
		\eIf{$\bar{x}^*\in S$}
		{$f_{opt} \leftarrow f(\bar{x}^*)$;}{$L \leftarrow \{P\}$;}
		
		\textbf{Step 2: Node Operations (Parallel B\&B)}
		
		\While{$L\neq \emptyset$}
		{
			Select a sublist $L_s$ of $L$ with at most $s$ nodes in $L_s$;
			
			Update $L\leftarrow L \setminus L_s$;
			
			\SetKwFor{For}{parallelfor}{do}{end}   
			\For{$P_i \in L_s$}
			{
				Solve $R(P_i)$ and get its solution $x^*$ and lower bound $l(P_i)$;		
				
				\SetKwIF{If}{ElseIf}{Else}{if}{then}{else if}{else}{end}
				\uIf{$R(P_i)$ is feasible and $l(P_i) < f_{opt}$}
				{
					\uIf {$x^* \in S$}
					{
						$x_{opt}\leftarrow x^*$; $f_{opt} \leftarrow l(P_i)$; 
					}
					\Else
					{
						\uIf {$f_{opt}-l(P_i)> \varepsilon$}
						{
							Run DCA for $(P_i^t)$ from $x^*$ to get its solution $\hat{x}^*$;
							
							\uIf {$\hat{x}^* \in S$ and $f_{opt} > f(\hat{x}^*)$}
							{
								$x_{opt}\leftarrow \hat{x}^*$; $f_{opt} \leftarrow f(\hat{x}^*)$; 
								
							}{}
						}
						\Else
						{
							Branch $P_i$ into two new problems $P_i^u$ and $P_i^d$;
							
							Update $L\leftarrow \{P_i^u , P_i^d\}$;
						}
					}
				}
			}
		}
	}
\end{algorithm}

\section{Experimental Results}\label{sec:experimental_result}
In this section, we present our experimental results for assessing the performance of the sentence compression model described above. 

Our sentence compression model is implemented in Python as a Natural Language Processing package, called `NLPTOOL' (actually supporting multi-language tokenization, tagging, parsing, automatic CFG grammar generation, and sentence compression), which implants NLTK 3.2.5\cite{software_NLTK} for creating parsing trees and Gurobi 7.5.2\cite{software_gurobi} for solving the linear relaxation problems $R(P_i)$ and the convex optimization subproblems in Step 2 of DCA. The PDCABB algorithm is implemented in C++ and invoked in python. The parallel computing part in PDCABB is realized by OpenMP. 

\subsection{F-score evaluation}

\modif{
We use a statistical approach called \emph{F-score} to evaluate the similarity between the compression computed by our algorithm and a standard compression provided by human. F-score is defined by : $$F_{\mu} = (\mu^2+1)\times \frac{P\times R}{\mu^2 \times P + R}$$
where $P$ and $R$ represent for precision rate and recall rate as:
	$$P = \frac{A}{A+C}, R = \frac{A}{A+B}$$
in which $A$ denotes for the number of words both in the compressed result and the standard result; $B$ is the number of words in the standard result but not in the compressed result; and $C$ counts the number of words in the compressed result but not in the standard result. The parameter $\mu$, called preference parameter, stands for the preference between precision rate and recall rate for evaluating the quality of the results. $F_{\mu}$ is a strictly monotonic function defined on $[0,+\infty[$ with $\displaystyle\lim_{\mu\to 0} F_{\mu}=P$ and $\displaystyle\lim_{\mu\to +\infty} F_{\mu}=R$. In our tests, we will use $F_1$ as F-score. Clearly, a bigger F-score indicates a better compression.}

\subsection{Numerical Results}
\begin{figure}[htb]
	\centering
	\caption{Box-plots for different models v.s. F-scores}
	\label{fig:boxplot01}
	\includegraphics[width=0.8\linewidth]{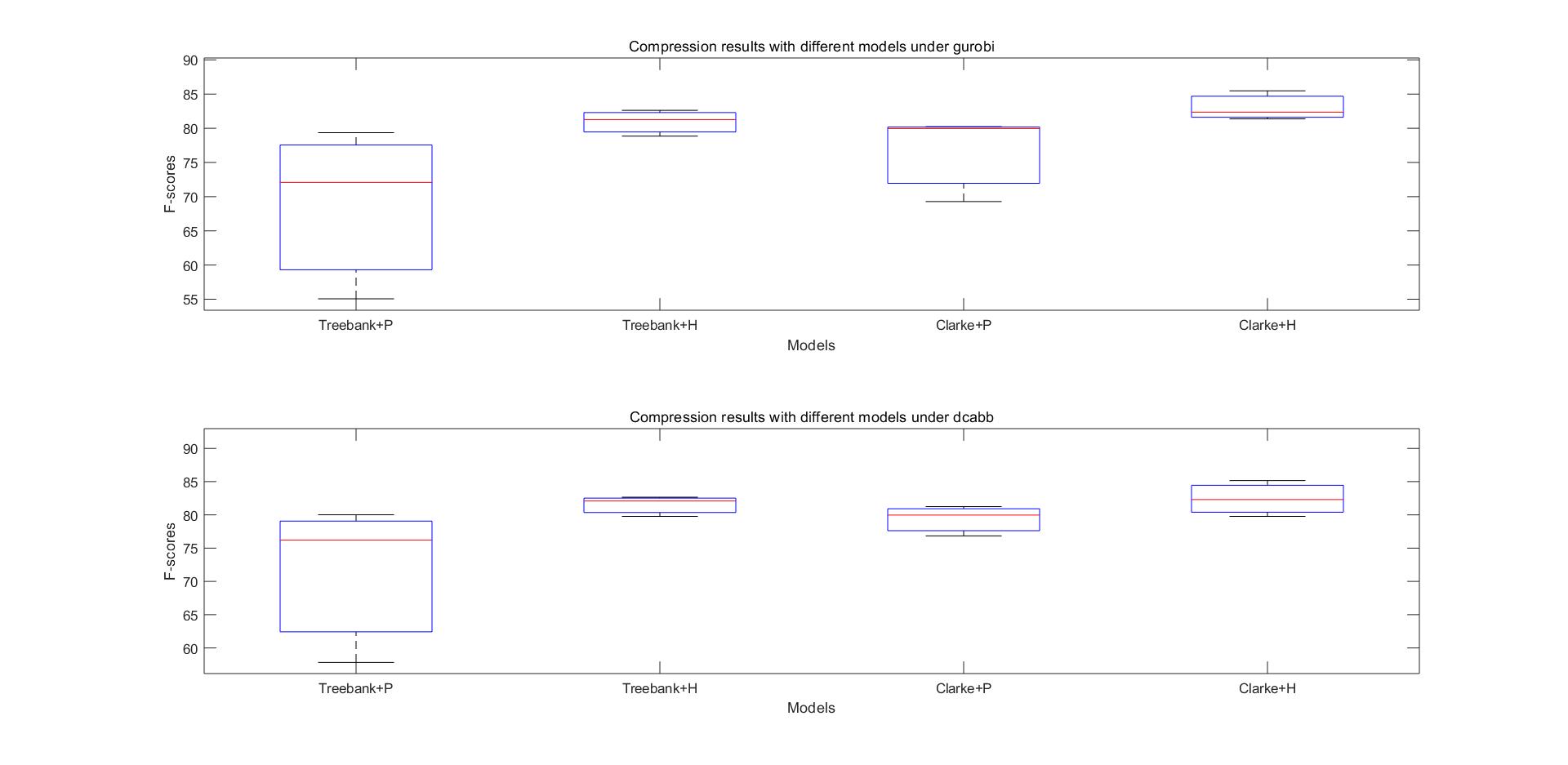}
\end{figure}

\modif{
	Table \ref{tab:results1} illustrates the compression result of $100$ sentences obtained by two ILP compression models: our new hybrid model (H) v.s. the probability model (P). Penn Treebank corpus (Treebank) provided in NLTK and CLwritten corpus (Clarke) provided in \cite{paper_clarke_2008} are used for sentence compression. We applied Kneser-Ney Smoothing for computing trigram probabilities. The compression rates \footnote{The compression rate is computed by the length of compression over the length of original sentence.} are given by $50\%$, $70\%$ and $90\%$. We compare the average solution time and the average F-score for these models solved by Gurobi and PDCABB. The experiments are performed on a laptop equipped with $2$ Intel i$5$-$6200$U $2.30$GHz CPU ($4$ cores) and $8$ GB RAM.
\vspace{-10pt}
\begin{table}[htb!]
	\caption{Compression results}\label{tab:results1}
	\centering
	\resizebox{300pt}{45pt}{
	\begin{tabular}[h]{|c|c|c|c|c|c|c|c|}	
		\hline
		\multirow{2}*{Corpus+Model} & \multirow{2}*{Solver} & \multicolumn{2}{c|}{$50 \%$ compression rate} & \multicolumn{2}{c|}{$70 \%$ compression rate} & \multicolumn{2}{c|}{$90 \%$ compression rate} \\
		\cline{3-8}
		& & F-score (\%) & Time (s) & F-score (\%) &  Time (s) &  F-score (\%) &  Time (s)\\
		\hline
		\multirow{2}*{Treebank+P} & Gurobi & 56.5 & 0.099 & 72.1 & 0.099 & 79.4 & 0.081 \\
		\cline{2-8}
		& PDCABB & {\bfseries 59.1} & 0.194 &{\bfseries 76.2} & 0.152 & {\bfseries 80.0} & 0.122\\
		\hline
		\multirow{2}*{Treebank+H} & Gurobi & 79.0 & 0.064 & 82.6 & 0.070 & 81.3 & 0.065\\
		\cline{2-8}
		& PDCABB & {\bfseries 79.9} & 0.096 & {\bfseries 82.7} & 0.171 & {\bfseries 82.1} &0.121\\
		\hline
		\multirow{2}*{Clarke+P} & Gurobi & 70.6 & 0.087 & \bfseries 80.2 & 0.087 & 80.0 & 0.071\\
		\cline{2-8}
		& PDCABB &{\bfseries 81.4} & 0.132 & 80.0 & 0.128 & {\bfseries 81.2} & 0.087\\
		\hline
		\multirow{2}*{Clarke+H} & Gurobi &  77.8 & 0.046 & \bfseries 85.5 & 0.052 & \bfseries 82.4 & 0.041\\
		\cline{2-8}
		& PDCABB & {\bfseries 79.9} & 0.081 & 85.2 & 0.116 & 82.3 & 0.082\\
		\hline
	\end{tabular}}
\end{table}

It can be observed that our hybrid model often provides better F-scores in average for all compression rates, while the computing time for both Gurobi and PDCABB are all very short within less than 0.2 seconds. We can also see that Gurobi and PDCABB provided different solutions since F-scores are different. This is due to the fact that branch-and-bound algorithm find only approximate global solutions when the gap between upper and lower bounds is small enough. Even both of the solvers provide global optimal solutions, these solutions could be also different since the global optimal solution for ILP could be not unique. However, the reliability of our judgment can be still guaranteed since these two algorithms provided very similar F-score results.

The box-plots given in Figure \ref{fig:boxplot01} demonstrates the variations of F-scores for different models with different corpora. We observed that our hybrid model (Treebank+H and Clarke+H) provided better F-scores in average and is more stable in variation, while the quality of the compressions given by probability model is worse and varies a lot. Moreover, the choice of corpora will affect the compression quality since the trigram probability depends on corpora. Therefore, in order to provide more reliable compressions, we have to choose the most related corpora to compute the trigram probabilities. 
}
\vspace{-25pt}
\vspace{-25pt}

\section{Conclusion and Perspectives}\label{sec:conclusion_perspectives}

\modif{We have proposed a hybrid sentence compression model ILP-PT based on the probability model and the parse tree model to guarantee the syntax correctness of the compressed sentence and save the main meaning. We use a DC programming approach PDCABB to solve our sentence compression model. Experimental results show that our new model and the solution algorithm can produce high quality compressed results within a short compression time.

Concerning on future works, we are very interested in designing a suitable recurrent neural network for sentence compression. With deep learning method, it is possible to classify automatically the sentence types and fundamental structures, it is also possible to distinguish the fixed collocation in a sentence and make these variables be remained or be deleted together. Researches in these directions will be reported subsequently.}

%

%
%
%

\end{document}